%% file: od-qat.tex
\documentclass[10pt,twocolumn,letterpaper]{article}

\usepackage[pagenumbers]{wacv} 

\usepackage{graphicx}
\usepackage{amsmath}
\usepackage{amssymb}
\usepackage{booktabs}
\usepackage[smallcaps]{glossaries}[=v4.46]

\usepackage[pagebackref,breaklinks,colorlinks]{hyperref}
\usepackage{tabularray}
\usepackage{mathtools}

\input{math_commands}

\input{macros}

\usepackage{colortbl}

\definecolor{myGray}{gray}{0.9}

\definecolor{Gray}{gray}{0.85}
\newcolumntype{z}{>{\columncolor{myGray}}c}
\newcolumntype{a}{>{\columncolor{myGrayy}}c}
\newcolumntype{b}{>{\columncolor{myGrayyy}}c}

\usepackage[capitalize]{cleveref}
\crefname{section}{Sec.}{Secs.}
\Crefname{section}{Section}{Sections}
\Crefname{table}{Table}{Tables}
\crefname{table}{Tab.}{Tabs.}

\def\confName{WACV}
\def\confYear{2024}

\begin{document}

\title{Reducing the Side-Effects of Oscillations in Training of Quantized  \\ YOLO Networks}

\author{%
  Kartik Gupta ~~~~~~~~~
  Akshay Asthana 
\\
Seeing Machines, Australia\\
{\tt\small \{kartik.gupta,akshay.asthana\}@seeingmachines.com}\\
}

\maketitle

\begin{abstract}
Quantized networks use less computational and memory resources and are suitable for deployment on edge devices. While quantization-aware training (\acrshort{QAT}) is a well-studied approach to quantize the networks at low precision, most research focuses on over-parameterized networks for classification with limited studies on popular and edge device friendly single-shot object detection and semantic segmentation methods like \yolo. Moreover, majority of \acrshort{QAT} methods rely on \gls{STE} approximation which suffers from an oscillation phenomenon resulting in sub-optimal network quantization. In this paper, we show that it is difficult to achieve extremely low precision (4-bit and lower) for efficient \yolo models even with SOTA \acrshort{QAT} methods due to oscillation issue and existing methods to overcome this problem are not effective on these models. To mitigate the effect of oscillation, we first propose Exponentially Moving Average (\exma{}) based update to the \acrshort{QAT} model. Further, we propose a simple \acrshort{QAT} correction method, namely \qc{}, that takes only a single epoch of training after standard \gls{QAT} procedure to correct the error induced by oscillating weights and activations resulting in a more accurate quantized model. With extensive evaluation on \coco dataset using various \yoloa and \yolob variants, we show that our correction method improves quantized \yolo networks consistently on both object detection and segmentation tasks at low-precision (4-bit and 3-bit).

\end{abstract}

\input{text/intro.tex}

\input{text/relatedwork.tex}

\input{text/prelim.tex}

\input{text/oscillation-in-yolo.tex}

\input{text/method}

\input{text/experiments}

\input{text/conclusion}

\onecolumn
\appendices

\input{text/appendix}

\twocolumn
{\small
\bibliographystyle{ieee_fullname}
\bibliography{od-qat}
}

\end{document}

%% file: math_commands.tex
\usepackage{amsmath,amsfonts,bm}
\usepackage{upgreek}

\def\figref#1{figure~\ref{#1}}

\def\secref#1{section~\ref{#1}}

\def\eqref#1{equation~\ref{#1}}

\def\1{\bm{1}}

\def\rva{{\mathbf{a}}}
\def\rvb{{\mathbf{b}}}

\def\rvh{{\mathbf{h}}}

\def\rvs{{\mathbf{s}}}

\def\rvw{{\mathbf{w}}}
\def\rvx{{\mathbf{x}}}
\def\rvy{{\mathbf{y}}}

\def\rmW{{\mathbf{W}}}

\DeclareMathAlphabet{\mathsfit}{\encodingdefault}{\sfdefault}{m}{sl}
\SetMathAlphabet{\mathsfit}{bold}{\encodingdefault}{\sfdefault}{bx}{n}

\DeclareMathOperator*{\argmin}{arg\,min}

\renewcommand\vec[1]{\mathbf{#1}}

\newcommand{\R}{\mathbb{R}}

\DeclareMathOperator{\BatchNorm}{BatchNorm}

\renewcommand\vec[1]{\mathbf{#1}}
\newcommand\vecq[1]{\vec{\widehat{#1}}}
\newcommand\vect[1]{\vec{\widetilde{#1}}}

\newcommand\ct[1]{#1}
\renewcommand\vec[1]{\mathbf{#1}}
\newcommand\tlossb[0]{\mathcal{L}}

\newcommand\bnbeta{\boldsymbol{\upbeta}}
\newcommand\bngamma{\boldsymbol{\upgamma}}

\newcommand\eopX[1]{\mathop{\mathbb{E}_{#1}}}

\DeclarePairedDelimiter\round{\lfloor}{\rceil}

\newcommand\clip[3]{\text{clip}\left(#1,#2,#3\right)}

\newcommand\softround[1]
{\text{softround}\left(#1\right)}



%% file: macros.tex
\usepackage{float}
\usepackage{mathtools}
\usepackage{framed}
\usepackage{pbox}
\usepackage{afterpage}
\usepackage{url}
\usepackage{array}
\usepackage{amsfonts}

\usepackage{multirow}
\usepackage{booktabs}
\usepackage{relsize}
\usepackage{xspace}
\usepackage{caption}
\usepackage{subcaption}

\newcommand{\ignore}[1]{}

\usepackage{framed}	
\usepackage{url}
\usepackage{enumerate}
\usepackage{color}

\usepackage{changepage}

\usepackage{algorithmic}
\usepackage{algorithm}

\usepackage{amsthm}
\theoremstyle{definition}
\theoremstyle{plain}

\newcommand{\SKIP}[1]{}

\renewcommand{\R}{\rm I\!R}

\newcommand{\calD}{\mathcal{D}}

\newcommand{\alllayers}{\{1,\ldots, \ell\}}

\usepackage{xcolor}

\usepackage{bbm}

\usepackage[page]{appendix}
\usepackage{chngcntr}
\usepackage{etoolbox}

\AtBeginEnvironment{subappendices}{%
\section*{Appendix}
\addcontentsline{toc}{section}{Appendix}
}

\usepackage{xspace}
\makeatletter
\DeclareRobustCommand\onedot{\futurelet\@let@token\@onedot}
\def\@onedot{\ifx\@let@token.\else.\null\fi\xspace}

\def\ie{\emph{i.e}\onedot}

\def\etal{\emph{et al}\onedot}
\makeatother

\usepackage{enumitem}

\def\figref#1{Fig.~\ref{#1}}

\def\secref#1{Sec.~\ref{#1}}
\def\eqref#1{Eq.~(\ref{#1})}

\def\tabref#1{Table~\ref{#1}}

\def\tabref#1{Table~\ref{#1}}

\glsdisablehyper
\newacronym{SGD}{sgd}{Stochastic Gradient Descent}
\newacronym{GD}{gd}{Gradient Descent}
\newacronym{DNN}{dnn}{Deep Neural Networks}
\newacronym{NN}{nn}{Neural Network}
\newacronym{fc}{fc}{fully-connected}
\newacronym{KL}{kl}{KL}
\newacronym{S}{s}{S}
\newacronym{LR}{lr}{LR}
\newacronym{STE}{ste}{Straight Through Estimator}
\newacronym{QN}{qn}{Quantization Networks}
\newacronym{QAT}{qat}{Quantization-Aware Training}
\newacronym{FC}{fc}{FC}
\newacronym{BN}{bn}{Batch Normalization}
\newacronym{EMA}{ema}{Exponential Moving Average}
\newacronym{PTQ}{ptq}{Post-training Quantization}
\newacronym{LSQ}{lsq}{Learnable Step-size Quantization}

\newacronym{QC}{qc}{Quantization Correction}
\newacronym{YOLO}{yolo}{You Only look Once}
\newacronym{YOLO5}{yolo5}{YOLO5}
\newacronym{YOLO7}{yolo7}{YOLO7}
\newacronym{ADAM}{adam}{ADAM}

\newcommand{\exma}{\acrshort{EMA}}
\newcommand{\qc}{\acrshort{QC}}

\newcommand{\coco}{{\small COCO}\xspace}
\newcommand{\yoloa}{{\small YOLO5}\xspace}
\newcommand{\yolo}{{\small YOLO}\xspace}
\newcommand{\yolob}{{\small YOLO7}\xspace}

\newcommand{\imagenet}{{ImageNet}}
\newcommand{\femnist}[1]{{\small FEMNIST}}

\usepackage{pifont}
\newcommand{\cmark}{\ding{51}} 
\newcommand{\xmark}{\ding{55}} 

\usepackage{wrapfig}

\newif\ifsupp
\suppfalse

\newif\ifarxiv
\arxivfalse

\newif\iffinal
\finalfalse



%% file: text/intro.tex
\section{Introduction}
Deep neural networks have achieved remarkable success in various applications, including image classification, object detection, and semantic segmentation. However, deploying these models on edge devices such as mobile phones, smart cameras, and drones poses a significant challenge due to their limited computational and memory resources. These devices typically have limited battery life, storage capacity, and processing power, making it challenging to execute complex neural networks.
To overcome these challenges, researchers have developed techniques for optimizing neural networks to reduce their computational and memory requirements while maintaining their accuracy. One such line of research is \gls{QAT}, which reduces the number of bits used to represent the network parameters, and activations resulting in smaller model sizes and faster inference times. Existing \acrshort{QAT} \cite{lsq, liu2022nonuniform, nagel2021white, krishnamoorthi} methods have made remarkable progress in quantizing neural networks at ultra-low precision with the effectiveness of {\em \acrfull{STE}} approximation still being a point of study. Previous works~\cite{Gong2019DSQ, quantization_nets} have proposed smooth approximation of rounding function to avoid the use \acrshort{STE} approximation but \acrshort{STE} is still considered to be the de-facto method for approximating gradient of quantization function during propagation due to its simplicity. Furthermore, recent works \cite{nagel2022overcoming, defossez2022differentiable} have shown oscillation issue affects quantization performance of efficient network architecture at low-precision due to \acrshort{STE} approximation in \acrshort{QAT}. 

Apart from that, the majority of \acrshort{QAT} literature focuses on image classification, and quantization performance achieved on such classification tasks does not necessarily translate onto downstream tasks such as object detection, and semantic segmentation. In this paper, we focus on the more challenging task of quantizing the single-shot efficient detection networks such as \yoloa{}~\cite{yolov5} and \yolob{}~\cite{wang2022yolov7} at low-precision (3-bits and 4-bits). Furthermore, we show that the oscillation issue is even more prevalent on these networks and the gap between full-precision and quantized performance is far from what is usually observed in \acrshort{QAT} literature. We also show that apart from latent weights, learnable scale factors for both weights and activations are also affected by the oscillation issue in \yolo{} models and latent weights around quantization boundaries are sometimes closer to optimality than quantization levels. This indicates that per-tensor quantization worsens the issue of oscillation.

To deal with the issues of oscillations in \yolo, we propose {\em Exponential Moving Average (\exma{})} in \acrshort{QAT}, that smoothens out the effect of oscillations and {\em Quantization Correction (\qc)}, that corrects the error induced due to oscillation after each quantized layer as a post-hoc step after performing \acrshort{QAT}. By mitigating side-effects of oscillations, these two methods in combination achieve state-of-the-art quantization results at 3-bit and 4-bit on \yoloa{} and \yolob{} for both object detection and semantic segmentation on extremely challenging \coco{} dataset. 

Below we summarize the contributions of this paper:
\begin{itemize}[leftmargin=*]
\item We show that quantization on most recent efficient \yolo{} models such as \yoloa{} and \yolob{} is extremely challenging even with state-of-the-art \acrshort{QAT} methods due to oscillation issue.
\item Our analysis finds that the oscillation phenomenon does not only affect latent weights but also affects the training of learnable scale factors for both weights and activations. 
\item We propose two simple methods namely \exma{} and \qc{}, that can be used in combination with any \acrshort{QAT} technique to reduce the side-effects of oscillations during \acrshort{QAT} on efficient networks. 
\item With extensive experiments on \coco{} dataset for both object detection and semantic segmentation tasks, we show that our methods in combination consistently improve quantized \yoloa{} and \yolob{} variants and establish a state-of-the-art at ultra-low precision (4-bits and 3-bits).
\end{itemize}

%% file: text/relatedwork.tex
\section{Related Work}
\paragraph{Quantization-Aware Training.}
In recent years, model quantization has been a topic of great interest in the deep learning community due to neural networks continuously scaling exponentially in terms of compute. Neural network quantization approaches can be broadly categorized into: \acrfull{PTQ} and \acrfull{QAT}. Though \acrshort{PTQ} \cite{dfq, bannerposttraining, nagel2021white, adaround} is faster and does not rely on whole training data, it yields significant performance degradation at low-bit precision. \acrshort{QAT} is the focus of our work, and it has been well-studied in literature \cite{Gong2019DSQ, pact2018, zhou2017incremental, lsq, liu2022nonuniform, EWGS}.

The \acrshort{STE} is a de-facto method for backpropagating through the non-differentiable rounding function in \acrshort{QAT}. The effectiveness of \acrshort{STE} has been the point of argument in recent literature. Ajanthan \etal \cite{pmlr-v130-ajanthan21a} proposed a mirror descent formulation for neural network quantization and established the connection between \acrshort{STE} approximation and mirror descent framework for constrained optimization. Lee \etal~\cite{EWGS} showed that the \acrshort{STE} leads to bias in gradients and proposed gradient scaling by the distance of latent weights from quantization boundaries. Gong \etal \cite{Gong2019DSQ} attempt to mitigate the issues caused by \acrshort{STE} for low-bit quantization by using differentiable hyperbolic tangent functions to simulate the rounding function in the backward pass. Similar to that, Yang \etal~\cite{quantization_nets} approximate the rounding function using smooth sigmoid functions to address the gradient bias in \acrshort{STE}. 

Recent works~\cite{defossez2022differentiable, nagel2022overcoming} have identified oscillation as a side-effect of \acrshort{STE} approximation during \acrshort{QAT}. Defossez \etal~\cite{defossez2022differentiable} proposed additive Gaussian noise to mimic the quantization noise and replace it as quantization operation during \acrshort{QAT} to prevent weight oscillations and biased gradients resulting from \acrshort{STE}. Nagel \etal \cite{nagel2022overcoming} also showed that weight oscillation seriously impacts \acrshort{QAT} performance, specifically on efficient networks comprising depth-wise convolutions due to \acrshort{STE} approximation of rounding function. They propose to constrain the latent weights to avoid oscillation by either regularizing them to their quantized states or by freezing them. Recently, Liu \etal~\cite{liu2023oscillation} studied the issue of oscillations on vision transformers and proposed fixed scale factors for weight quantization and query-key re-parameterization to mitigate the negative influence of oscillation. However, their proposed approach is specifically targeted at solving oscillation phenomenon on transformers architecture. 

\paragraph{Quantization of  Object Detectors.}
The majority of existing neural network quantization literature focuses on the image classification task rather than real-world downstream tasks such as object detection or semantic segmentation. 
Some recent works do explore the quantization of object detectors to improve the efficiency of these models. Jacob \etal~\cite{jacob2018cvpr} proposed a quantization scheme using integer-only arithmetic and performing object detection on \coco{} dataset but the approach is only effective for 8-bit quantization. Li et al.~\cite{li2019fully} observed training instability during the quantized fine-tuning of RetinaNet and propose mutliple solutions specific to RetinaNet architecture. These solutions are not applicable on more efficient object detectors like \yolo{}. Ding \etal~\cite{ding2019req} proposed ADMM based weight quantization framework for \yolo{3} but do not address the issue of oscillations and activation quantization.  Due to the discrete nature of quantization functions, gradient estimates are known to be noisy, which affects \acrfull{SGD} updates. To overcome that, Zhuang \etal~\cite{zhuang2020training} proposed to utilize a full precision auxiliary module to enable stable training of quantized object detectors. Furthermore, Zhuang \etal~\cite{chen2021aqd} proposed multi-level \acrfull{BN} to accurately calculate batch statistics for each pyramid level in RetinaNet~\cite{lin2017focal} and FCOS~\cite{tian2019fcos}. The proposed approach is specific to pyramid-level architecture in RetinaNet and FCOS that share \acrshort{BN} layers at different levels of the pyramid, but not applicable on more recent efficient SOTA \yolo{} 
models. In this paper, we identify the gap in the recent literature on quantized object detection and introduce state-of-the-art quantized object detectors using \yoloa{}~\cite{yolov5} and \yolob{}\cite{wang2022yolov7} variants.

%% file: text/prelim.tex
\label{sec:osc-qat-toy}
\begin{figure*}[ht]
     \centering
     \begin{subfigure}[b]{0.324\textwidth}
         \centering
         \includegraphics[width=0.99\textwidth]{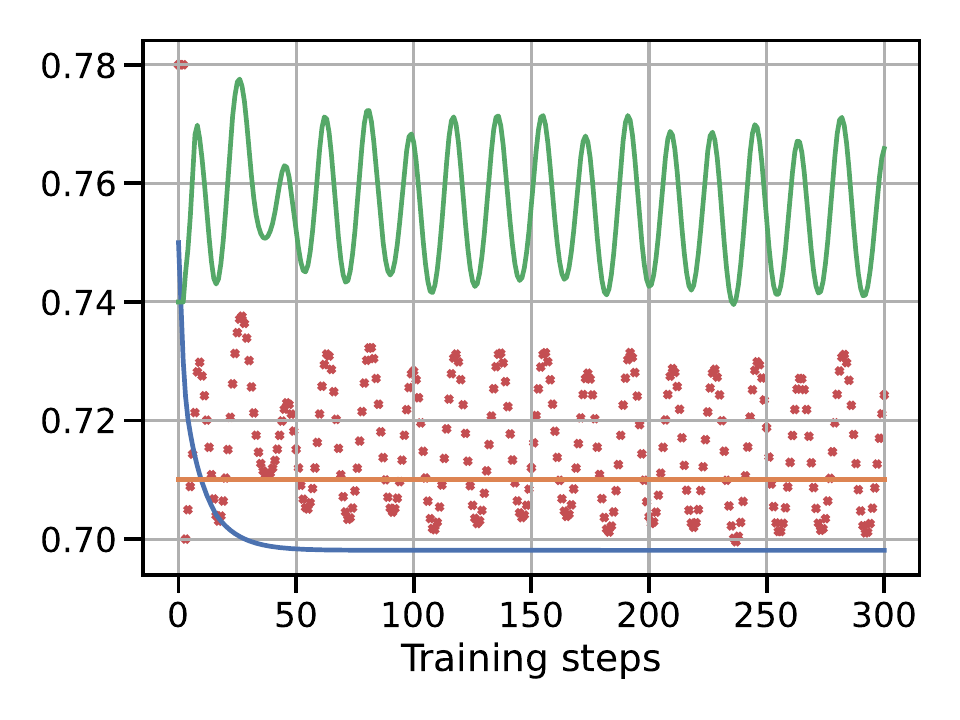}
         \caption{}
         \label{fig:toy_w1}
     \end{subfigure}
     \begin{subfigure}[b]{0.324\textwidth}
         \centering
         \includegraphics[width=0.99\textwidth]{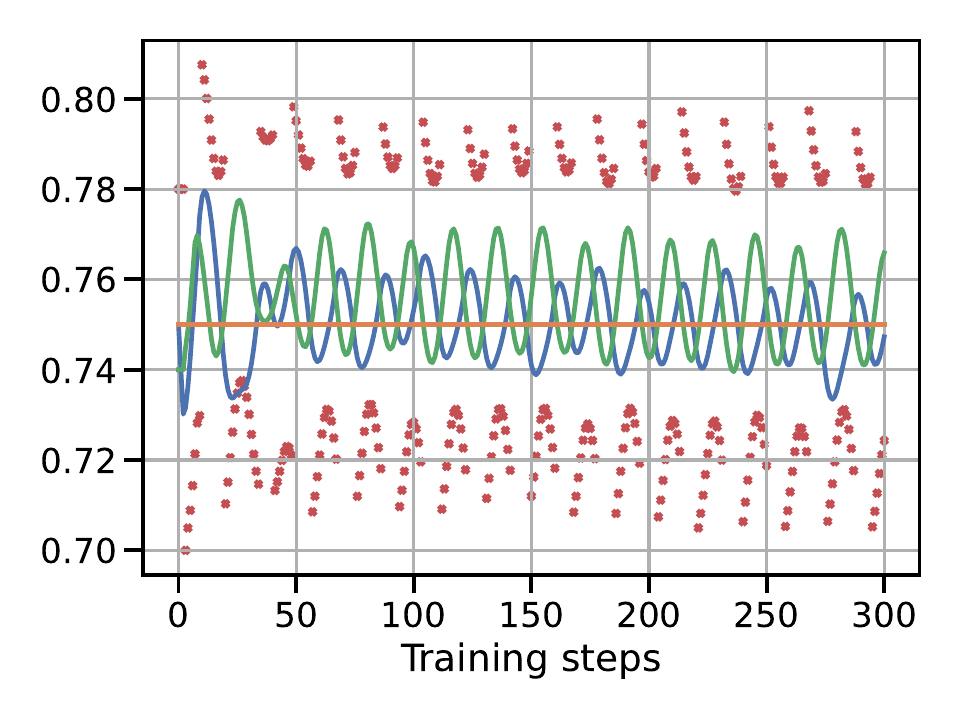}
         \caption{}
         \label{fig:toy_w2}
     \end{subfigure}
     \begin{subfigure}[b]{0.324\textwidth}
         \centering
         \includegraphics[width=0.99\textwidth]{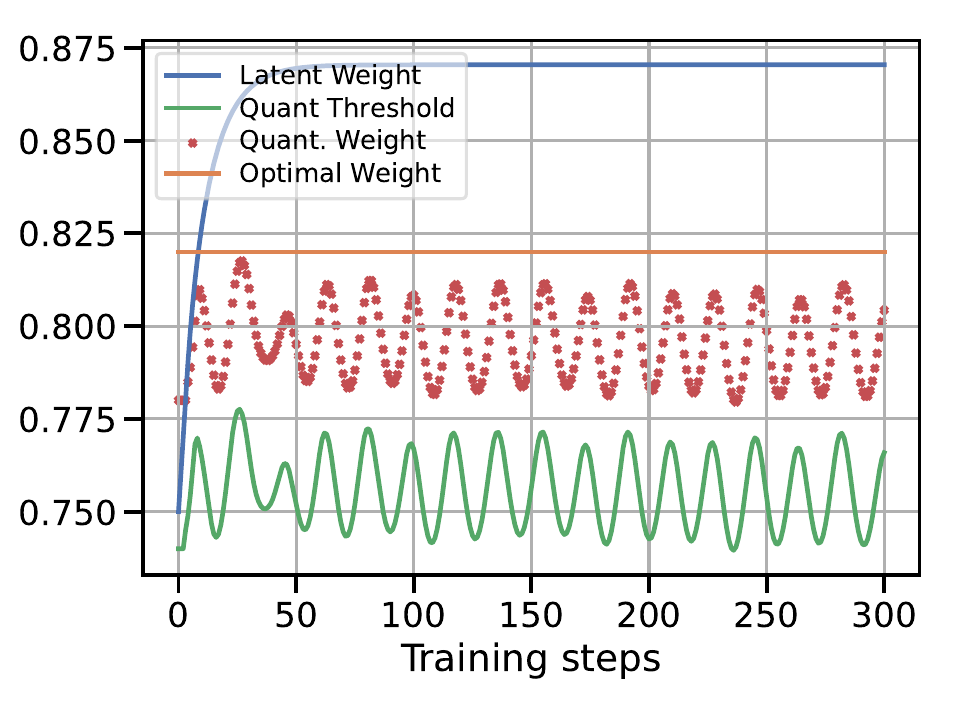}
         \caption{}
         \label{fig:toy_w3}
     \end{subfigure}
     \vspace{-2ex}
        \caption{\em A toy 3D regression problem to demonstrate the oscillation issue in weight and activation quantization. (a),(b) and (c) Trajectory of different weights during the optimization. Here, it can be seen that oscillation not only affects the latent weights but also the learnable scale factors. Here, "quantization threshold" refers to the quantization boundary in the latent space.} 
        \label{fig:oscillation_toy}
    \vspace{-3ex}
\end{figure*}
\section{Preliminaries}
\label{sec:prelim}
Here we provide a brief background on the quantization-aware training (\acrshort{QAT}) and introduce the issue of oscillations in \acrshort{QAT} using a small toy example.
\begin{figure}
\begin{center}
\includegraphics[width=0.70\linewidth]{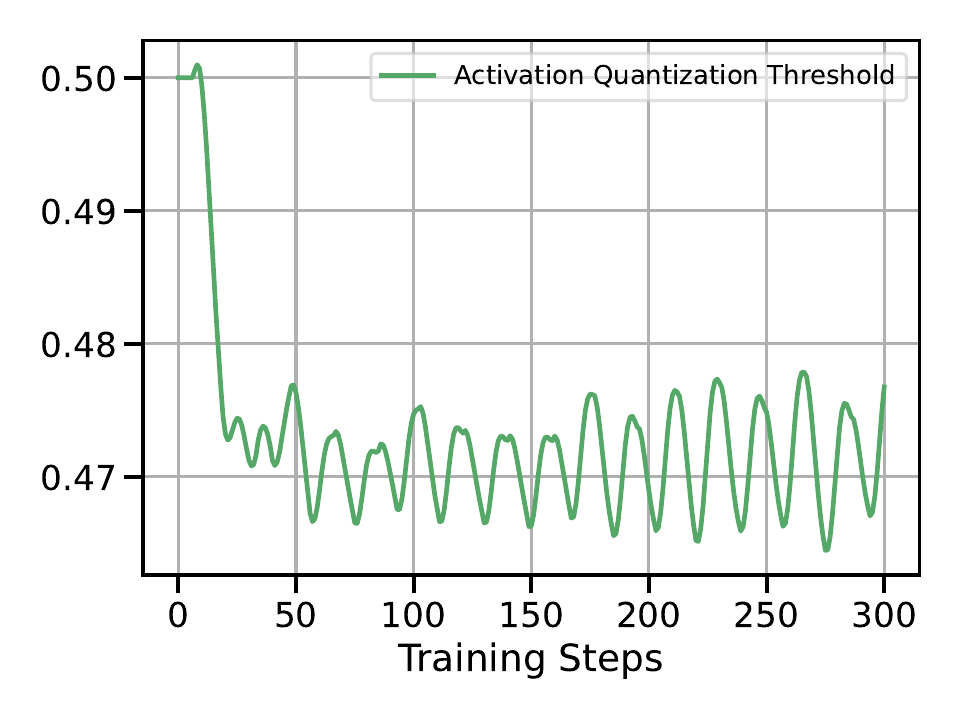}	
\end{center}
\vspace{-4ex}
\caption{\em Trajectory of activation quantization threshold during training for same toy example as in \figref{fig:oscillation_toy}. Even the scale factors for activation quantization oscillate during the optimization.}
 \label{fig:toy_x}
\end{figure}
\subsection{Quantization-aware Training 
(\acrshort{QAT})}
\label{sec:prelim-qat}
Quantization-aware training can be achieved by simulating the quantized computational operations during the training of the neural networks. The forward pass of the neural network is encapsulated with a quantization function $q(\cdot)$ that converts full-precision weights and activations into quantized weights and activations. It takes input vector $\vec{w}$ and returns quantized output $\vecq{w}$ given by:
\begin{equation}
    \label{eq:quantization_formulation}
    \vecq{w} = q(\vec{w}; s, u, v) = \ct{s} \cdot \clip{\round*{\frac{\vec{w}}{\ct{s}}}}{\ct{u}}{\ct{v}}, 
\end{equation} 

where $\round*{\cdot}$ is the \textit{round-to-nearest} operator, $\clip{\cdot}{b}{c}$ is a clipping function with lower and upper bounds $b$ and $c$, respectively, $\ct{s}$ is a quantization scale factor. Here, $\ct{u}$ and $\ct{v}$ denote the minimum and maximum range after the quantization. The scale factor $s$ can be learned \cite{lsq, lsq+} during quantization-aware training through backpropagation by approximating the gradient of the rounding operator with \acrshort{STE}. The original full-precision weights $\vec{w}$ are commonly referred to as \textit{latent weights} and gradient descent is performed only on the latent weights for the update. During inference, the quantized weights $\vecq{w}$ are used to compute the convolutional or dense layer output.

Due to the non-differentiability of the quantization function, it is non-trivial to backpropagate through the neural network embedded with such an operation. To this purpose, a commonly used technique for alleviating this issue involves using the straight-through estimator (\acrshort{STE}) \cite{bengio2013estimating, hinton2012ste}. The basic idea of \acrshort{STE} is to approximate the gradient of the rounding operator as 1 within the quantization limits. 

\subsection{Oscillations in \acrshort{QAT}}

\begin{figure*}[ht]
     \centering
     \begin{subfigure}[b]{0.32\textwidth}
         \centering
         \includegraphics[width=0.99\textwidth]{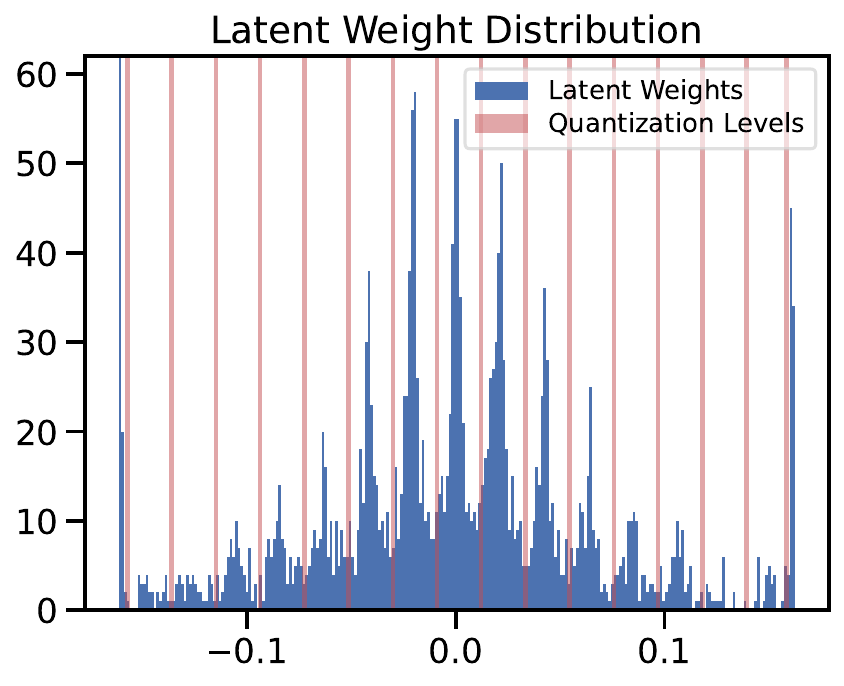}
         \caption{}
         \label{fig:latent_yolo}
     \end{subfigure}
     \begin{subfigure}[b]{0.33\textwidth}
         \centering
         \includegraphics[width=1.05\textwidth]{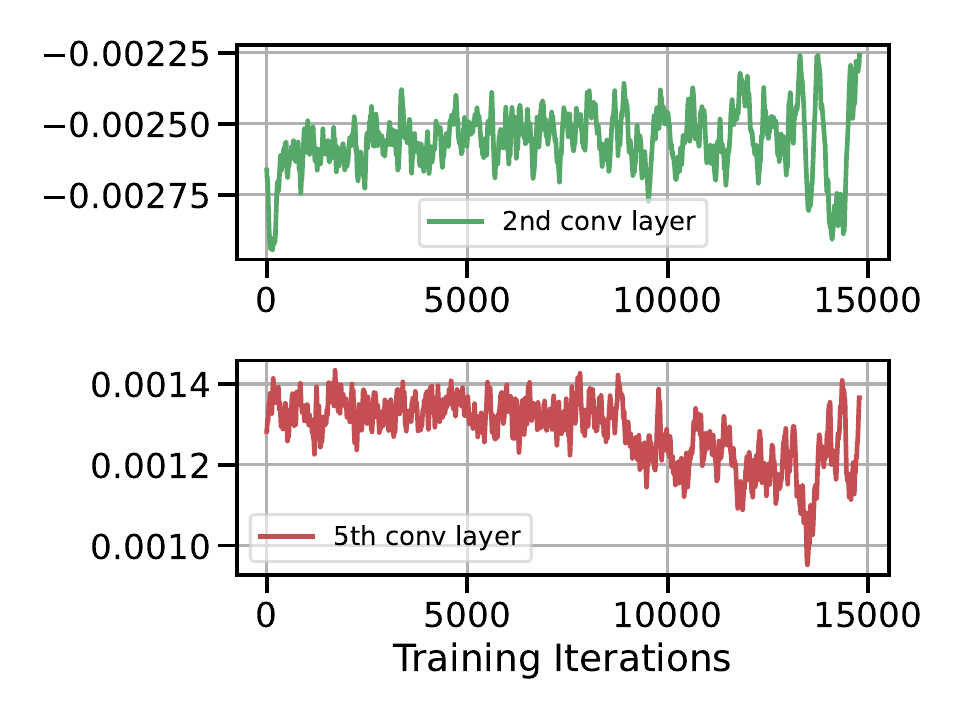}
         \caption{}
         \label{fig:scalefactor_wt_yolo}
     \end{subfigure}
     \begin{subfigure}[b]{0.33\textwidth}
         \centering
         \includegraphics[width=0.99\textwidth]{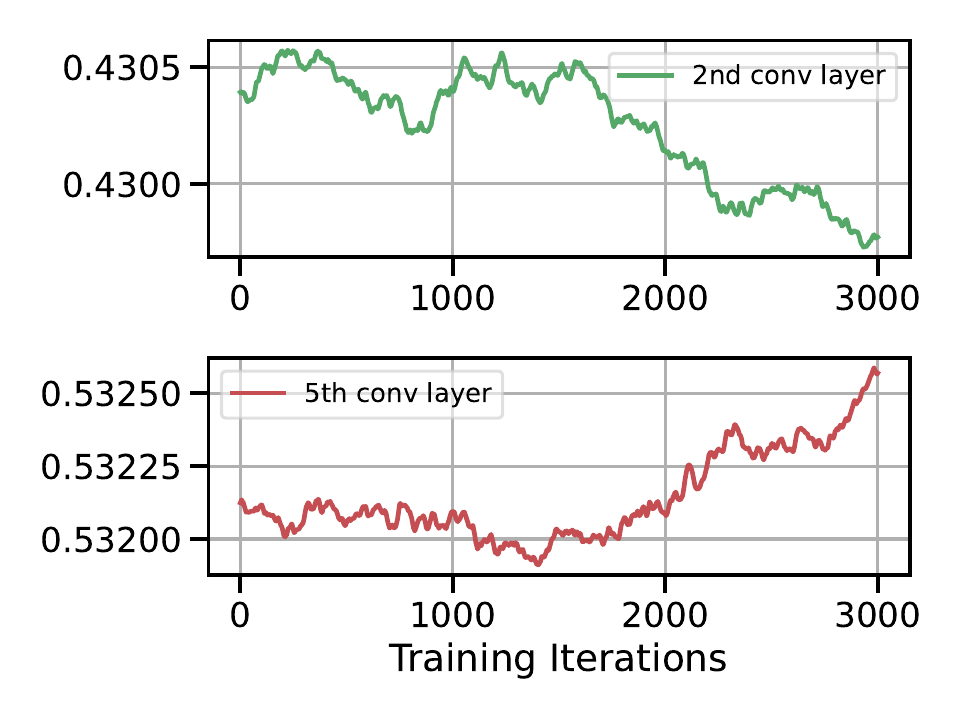}
         \caption{}
         \label{fig:scalefactor_act_yolo}
     \end{subfigure}
    \vspace{-2ex}
        \caption{\em Oscillation issue in \yoloa{}-n variant trained on \coco{} dataset at 4-bit precision using \acrshort{LSQ}~\cite{lsq}. (a) Latent weight distributions of $5^{th}$ convolution layer in \yoloa{}-n, and Scale factors for (b) weight quantization during the last 15K iterations, (c) Scale factors for activation quantization during the last 3K iterations of QAT in YOLOv5-n. Here, it can be clearly observed that latent weights peak exactly at quantization thresholds. Also, both quantization scale factors of weights and activations are not stable even until the end of training.}
        \label{fig:oscillation_yolov5}
    \vspace{-3ex}
\end{figure*}
\label{sec:oscillations}
Recent works~\cite{defossez2022differentiable,nagel2022overcoming} observed the oscillation phenomenon as a side-effect of quantization-aware training with \acrshort{STE} approximation. Due to \acrshort{STE} approximation that passes the gradient through the quantization function, the latent weights oscillate around the quantization threshold.

We illustrate the oscillation phenomenon in \gls{STE} based \acrshort{QAT} using a 3D toy regression problem with both weights and input quantization at 1-bit. Here, we optimize a latent weight vector $\vec{w}$ and scale factors $s_\vec{w}, s_\vec{x}$ for weights and activations respectively with the following objective: 
\begin{equation}
    \argmin_{\vec{w}, s_\vec{w}, s_\vec{x}}\tlossb(\vec{w}, s_\vec{w}, s_\vec{x}) = \eopX{\vec{x}\sim U}{ \|\vec{x}\vec{w}_* - q(\vec{x}, s_\vec{x})q(\vec{w}, s_\vec{w})\|}
    \label{eq:toy_regression_optim}
\end{equation}

Here, $\vec{w}_*$ refers to the optimal ground truth value and $q(\cdot)$ is the quantization function defined in \eqref{eq:quantization_formulation}. We randomly sample data vector $\vec{x}$ from a uniform distribution $U$ within range $[0,1)$. The oscillation behavior during the optimization is shown in \figref{fig:oscillation_toy}. Note that the quantized weights oscillate around the quantization threshold instead of converging close to the optimal value in \figref{fig:toy_w1}, \figref{fig:toy_w2}, and \figref{fig:toy_w3}. Though the underlying reason for oscillation of quantized weights in \figref{fig:toy_w2} is the oscillation of latent weights, oscillation of quantized weights in \figref{fig:toy_w1} and \figref{fig:toy_w3} happens due to oscillation of learnable scale factors. Furthermore, this oscillation behavior is not just restricted to quantized weights but also impacts the quantized activations as can be observed in \figref{fig:toy_x}. In the case of activation quantization, scale factors for activations oscillate as well and this can lead to further performance degradation of quantized models due to sub-optimal minima. Here, we would like to point out that previous work~\cite{nagel2022overcoming} only accounts for the issue of oscillation of latent weights, whereas we show oscillations in learnable scale factors can lead to performance degradation during \acrshort{QAT}.

%% file: text/oscillation-in-yolo.tex
\section{Side-effects of Oscillations in \yolo}
\label{sec:yolo-sideeffects}
The issue of oscillating weights and activations in quantization-aware training is not just restricted to a small toy problem but occurs in practice on \acrshort{YOLO} networks \cite{yolov5, wang2022yolov7} trained for the task of object detection and semantic segmentation. This leads to a significant loss in the accuracy of quantized \acrshort{YOLO} models. In this section, we show the evidence of oscillation issue prevalent in weights and activations of quantized \acrshort{YOLO} networks and how learning a single scale factor for each tensor is the underlying cause for sub-optimal latent weights.
\subsection{Oscillation Issue in YOLO networks}
\label{sec:yolo-oscillation}

We demonstrate the oscillation issue in \yolo networks using the latent weight distribution of $5^{th}$ layer in \yoloa-n variant \cite{yolov5} trained with 4-bit quantization on \coco dataset \cite{lin2014microsoft} using \acrfull{LSQ} \cite{lsq} with \gls{STE} approximation in \figref{fig:latent_yolo}. Most of the latent weights lie in between the quantization levels and the peaks of distribution lie on quantization thresholds rather than the quantization levels itself. Since most of the latent weights lie around quantization thresholds, they tend to keep switching their quantization state even at the end of the training as also shown in \cite{nagel2022overcoming}. Further, we plot the learnable scale factors used to quantize weights as well as activations in \figref{fig:scalefactor_wt_yolo} and \figref{fig:scalefactor_act_yolo} respectively. The quantization scale factors remain unstable even until the end of quantization-aware training. The oscillation issue does not only affect the latent weights but also affects the scale factors of both weights and activations. This leads to a sub-optimal quantization state of both weights and activations of the final QAT model. Here, we would like to highlight that previous work~\cite{nagel2022overcoming} observed the issue of oscillations in latent weights but here we show that oscillation also affects quantization scale factors corresponding to weights and activations.

\input{tables/softround-yolo}
\subsection{Analysis using threshold-based Soft-Rounding}

\label{sec:yolo-softround}
Oscillation dampening~\cite{nagel2022overcoming} was introduced as one of the techniques to reduce the side-effect of oscillations. This method in essence regularizes the latent weights such that the distribution of latent weights and quantization weights overlap each other. We step away from that and look at the optimality of latent weights in their un-quantized state. For this analysis, we modify the original quantization to deduce a soft-rounding function that allows quantizing the weights closer to quantization levels and leave the latent weights around the quantization threshold in their latent state. We describe our soft-rounding function $q^*$ that can be used to softly round weights or activations using a threshold $k$ as 
\begin{align}
    \label{eq:softrounding_formulation}
    \vec{w^*} = q^*(\vec{w}; s, u, v) = \ct{s} \cdot \clip{\softround{\frac{\vec{w}}{\ct{s}}}}{\ct{u}}{\ct{v}},    
\end{align} 
\vspace{-2ex}
\begin{equation}
    \mbox{where}\quad \softround{x} = \begin{cases}
    \round{x}  & \text{if }  |x-\round{x}| \leq k \\
    x   & \text{otherwise. } 
    \end{cases} \nonumber
\end{equation}

This soft-rounding function can be used on both weights and activations to evaluate whether latent weights and activations stuck at the quantization thresholds (see \figref{fig:latent_yolo}) are already closer to their optimal state. We replace the quantization function described in \eqref{eq:quantization_formulation} with this soft-rounding function (at $k=0.45$) in all the quantized layers of pre-trained quantized \yolo{} models and evaluate the performance on \coco{} dataset and the results are presented in \tabref{tab:softround-yolo5}. Surprisingly, we found that the weights or activations in their latent state produce equivalent or better performance than the ones in the quantized state. This indicates that latent weights oscillate around the quantization boundaries partly because not all weights or activations within a tensor can be quantized with the same single scale factor as in the case of per-tensor quantization. If a single scale factor per-tensor is chosen, some weights can never reach optimality due to the limitation of per-tensor quantization.

%% file: tables/softround-yolo.tex
\begin{table}
    \small
    \centering
    \caption{\em Comparison between hard-rounding and soft-rounding (k=0.45). Here, soft-rounding outperforms hard-rounding function which is actually even used during \acrshort{QAT}.}
\vspace{-2ex}
\begin{tabular}{l|ccc}
\toprule
Method                         & \#-bits & \yoloa-n & 
\yoloa-s \\
\midrule
Full precision                 & FP      & 28.0       & 37.4     \\
\midrule
\multirow{2}{*}{Hard Rounding} & 4-bit       & 20.6    & 32.5    \\

                               & 3-bit        & 15.2     & 27.2     \\
                \midrule
\multirow{2}{*}{Soft Rounding} & \cellcolor{myGray} 4-bit        & \cellcolor{myGray}\textbf{21.2}     & \cellcolor{myGray}\textbf{32.6}     \\
                               & \cellcolor{myGray} 3-bit        & \cellcolor{myGray} \textbf{16.2}     & \cellcolor{myGray} \textbf{27.7}    
\\
\bottomrule
\end{tabular}
\vspace{-2ex}
\label{tab:softround-yolo5}
\end{table}

%% file: text/method.tex
\section{Approach}
\label{sec:approach}
In this section, we first provide the notations and formulate the quantization-aware training optimization problem with learnable scale factors. Then, we introduce our two simple methods to deal with side-effects of error induced due to oscillation in network parameters during \acrshort{QAT}. 
\vspace{-2ex}

\paragraph{Problem setup.} For notational convenience, we consider a fully-connected neural network with weights ${\rmW^l\in\R^{N_l\times N_{l-1}}}$, biases $\rvb^l\in\R^{N_{l-1}}$, pre-activations $\rvh^l\in\R^{N_l}$, and post-activations $\rva^l\in\R^{N_l}$, for $l \in \alllayers$ up to $K$ layers. Then, the feed-forward dynamics of the neural network with simulated quantization can be formulated as:
\begin{align}
\label{eq:forward-qat}
\rva^l = \phi(\BatchNorm(\rvh^l))\ ,\qquad \rvh^l = \widehat\rmW^l\widehat\rva^{l-1} + \rvb^l\ ,\\
\mbox{where}\qquad \widehat\rmW^l = q(\rmW^l, s_{\rmW}^l), \qquad  \widehat\rva^{l-1} = q(\rva^{l-1}, s_{\rva}^{l-1}).\nonumber
\end{align}
Here $\phi: \R \to\R$ is an elementwise nonlinearity, and the input is denoted by $\rva^0=\rvx^0\in\R^N$. We denote quantized weights and activations by $\widehat\rmW^l$ and $\widehat\rva^l$ respectively and use $s_\rmW^l$, $s_\rva^{l}$ to represent their respective quantization scale factors. For simplicity of notation, we further also express network weight parameters corresponding to all the layers as $\mathcal{W}=\{\rmW^i\}_{i=1}^\ell$. Similarly, we represent the vector corresponding to all scale factors for quantization of various weight tensors as $\rvs_{\rmW} = \{s_{\rmW}^i\}_{i=1}^\ell$ and scale factors for quantization of activation tensors as $\rvs_{\rva} = \{s_{\rva}^i\}_{i=1}^\ell$. Given dataset $\calD = \{\rvx_i, \rvy_i\}_{i=1}^n$, the typical neural network optimization problem for quantization-aware training with learnable scale factors can then be formulated as:
\begin{equation}\label{eq:qat_nn}
    \argmin_{\mathcal{W,\rvs_{\rvw}, \rvs_{\rva}}}L(\mathcal{W}, \rvs_{\rvw}, \rvs_{\rva};\calD).
\end{equation}
We now explain our two simple techniques to overcome the issue of oscillating weights and scale factors and compensate for sub-optimal latent weights stuck at the quantization boundaries.
\subsection{Exponential Moving Average (\exma) to Smoothen effect of Oscillations}
\label{sec:ema}

Weight averaging of multiple local minima by using multiple model checkpoints attained at cyclic learning rates with restarts, has been shown \cite{izmailov2018averaging,tarvainen2017mean} to lead to better generalization and wider minima. The idea of using weight averaging for final model inference was earliest suggested in \cite{polyak1992acceleration}. Later, the semi-supervised learning method \cite{tarvainen2017mean} and self-supervised learning methods \cite{grill2020bootstrap} utilized exponential moving average of weights to learn in a knowledge distillation manner.

To overcome the oscillating weights and quantization scale factors due to \acrshort{STE} approximation, we propose exponential moving average of latent weights and scale factors for both weights and activations during the optimization. \acrshort{STE} approximation approach leads to latent weights moving around the quantization boundary which leads to constantly changing latent weight states. Exponential moving average can take into account model weights at the last several steps of training and smoothen out the oscillation behavior and come up with the best possible latent state for oscillating weights. The final quantized state inference can be done using \exma{} weights and quantization parameters instead of the latest update.  

For trainable network weight parameters $\rmW_{(t)}^l$ at layer $l$, we can compute the corresponding exponentially moving average weights $\rmW_{(t)}^{\prime\,l}$ at $t$ iteration as:
\begin{align}
{\rmW}_{(t)}^{\prime\,l} = \alpha \cdot \rmW_{(t-1)}^{\prime\,l}  + (1 - \alpha) \cdot \rmW_{(t)}^{l} .
\end{align}
Similarly, we can also calculate the exponential moving average scale factors for both weights and activations as: 
\begin{align}
    \rvs_{\rmW\,(t)}^{\prime} &= \alpha \cdot \rvs_{\rmW\,(t-1)}^{\prime} + (1-\alpha) \cdot \rvs_{\rmW\,(t)} \\
    \rvs_{\rva\,(t)}^{\prime} &= \alpha \cdot \rvs_{\rva\,(t-1)}^{\prime} + (1-\alpha)\cdot \rvs_{\rva\,(t)}.
\end{align}

Here, $\alpha$ is used as a decay parameter that can be tuned to account for approximately $1/(1-\alpha)$ last \acrshort{SGD} updates to achieve the \exma{} model. We keep the decay parameter $\alpha$ to be $0$ at the start of the \acrshort{QAT} procedure to enable larger updates at the start of the training. The \exma{} parameters are updated after end of every iteration of the training update and thus do not require backpropagation. It is important to note here that oscillation dampening and iterative freezing proposed in \cite{nagel2022overcoming}, are only applicable on weights but oscillation issue due to scale factors is even present in activation quantization as shown in \secref{sec:osc-qat-toy}. To overcome that issue, \exma{} on scale factors of activations can tackle it more appropriately. Furthermore, we would also highlight here that other non-trainable parameters such \gls{BN} statistics in deep neural networks already utilize exponential moving average can improve the unstable \acrshort{BN} statistics if the momentum value is chosen appropriately. Recent methods \cite{nagel2022overcoming}, suggested that \acrshort{BN} Statistics re-estimation enables improvement in the corrupted \acrshort{BN} statistics occurring due to oscillation of latent weights. We would like to mention that corrupted \acrshort{BN} statistics is not the only reason for performance degradation resulting from oscillations in \acrshort{QAT}. Nevertheless, our \exma{} models yield stable updates to both latent weight, activations, and their respective scale factors.

\subsection{Post-hoc Quantization Correction (\qc)}
\label{sec:qat-correction}

In this section, we propose a simple correction step that can be performed in a post-hoc manner after quantization-aware training to overcome the error induced by oscillating latent weights and scale factors. As shown empirically in the \secref{sec:yolo-sideeffects}, oscillation results in the majority of latent weights hanging at the quantization boundaries. We have further shown in \secref{sec:yolo-softround} that latent weights hanging at quantization boundaries are already closer to optimality than their nearest quantization levels as the soft-rounded (based on \eqref{eq:softrounding_formulation}) \yolo{} models yield better performance than quantized ones. The oscillation of scale factors mainly happens due to ineffective quantization with a single quantization scale factor per-tensor apart from the bias of STE approximation. Intuitively, different regions in the tensor might require different scale factors for an accurate quantized approximation. 

Our post-hoc correction quantization step simply transforms the pre-activations $\rvh^l\in\R^{N_l}$ of all $l$ layers using an affine function, to compensate for error induced in matrix multiplications due to oscillations during the quantization-aware training. We can now formulate the modified feed-forward dynamics of the quantized neural network for the post-hoc error correction step as:
\begin{align}
\label{eq:forward-qat-correction}
 \vect{h}^l = \bngamma^l \cdot \rvh^l  + \bnbeta^l\  ,\quad \rvh^l = \widehat\rmW^l\widehat\rva^{l-1} + \rvb^l\ , \\
 \rva^l = \phi(\BatchNorm(\vect{h}^l))\ 
\end{align}
Here, $\vect{h}^l$ denotes the modified pre-activations after the affine transformation in layer $l$. Also, for layer $l$ we represent the affine function with scale correction parameters $\bngamma^l\in\R^{N_l}$ and shift correction parameters $\bnbeta^l\in\R^{N_l}$. For simplicity of notation, we further express a set of all correction scale parameters with $\mathcal{G} = \{\bngamma^i\}_{i=1}^\ell$ and correction shift parameters with $\mathcal{B} = \{\bnbeta^i\}_{i=1}^\ell$. We initialize these correction parameters as identity transformation. We then optimize for these correction parameters via backpropagation starting from a pre-trained QAT model with the following objective:
\begin{equation}\label{eq:qat-corr-nn}
\argmin_{\mathcal{G},\mathcal{B}}L(\mathcal{W}, \mathcal{G}, \mathcal{B}, \rvs_{\rvw}, \rvs_{\rva};\calD_c).
\end{equation}
We train these correction parameters on a small calibration set $\calD_c$, which is also part of the training set. Notice that, for a typical convolutional layer these correction factors will have dimensions the same as the number of output channels after the convolution operation. We would like to highlight that these extra set of correction parameters can be absorbed in \acrfull{BN} trainable parameters generally succeeding a convolution layer and do not result in an extra computational load on the hardware. It is important to note that our correction step is different from the \acrshort{BN} re-estimation step \cite{nagel2022overcoming} where \acrshort{BN} statistics are re-estimated on the dataset after \acrshort{QAT}. \acrshort{BN} re-estimation cannot recover from quantization error accumulated in forward propagation of quantized neural network, unlike our post-hoc correction step. In fact, re-estimating \acrshort{BN} statistics is not required since the exponential moving average in \acrshort{BN} statistics can enable a stable state of statistics if the momentum value is chosen appropriately. Furthermore, these correction parameters can also be stored as quantization scale factors by converting per-tensor quantization to per-channel quantization. The conversion from per-tensor quantization to per-channel quantization is a natural outcome of batch normalization folding \cite{whitepaper} where batch normalization parameters are folded into the quantization scale factors of weights to get rid of \acrshort{BN} at inference. Previous work~\cite{nagel2022overcoming} on weight oscillations in \acrshort{QAT} only evaluate their quantization-aware training mechanism on per-tensor quantization but still keep the \acrshort{BN} layers intact in train mode during the training. Their main motivation for oscillation avoidance is to get rid of corrupted \acrshort{BN} statistics during the training. However, general practise~\cite{nagel2021white} to support per-tensor quantization is to fold \acrshort{BN} parameters before \acrshort{QAT}.

%% file: text/experiments.tex
\section{Experiments}
\input{tables/yolo-qat-ours2}

In this section, we evaluate the effectiveness of our proposed \exma{} and \qc{} mechanisms to deal with side-effects of oscillations during \acrshort{QAT} on various \yoloa{} and \yolob{} variants on \coco{} dataset \cite{lin2014microsoft}. In all the experiments, we perform both weights and activation quantization. We present state-of-the-art results for low-precision (4-bit and 3-bit) on all \yoloa{} and \yolob{} variants for object detection. We also compare our method against standard baselines such as \acrshort{LSQ}~\cite{lsq} and Oscillation dampening~\cite{nagel2022overcoming}. We also perform some ablation studies to reflect the improvement of our method in comparison to per-channel quantization. Finally, we establish a state-of-the-art quantized \yoloa{} model on the task of semantic segmentation using \coco{} dataset. In summary, our results establish new state-of-the-art for quantized \yoloa{} and \yolob{} at low-bit precision while outperforming comparable baselines. 

\paragraph{Experimental Setup.} 
Similar to \cite{nagel2022overcoming}, we apply \acrshort{LSQ}~\cite{lsq} based weight and activation quantization. Since object detection is a complex downstream task and quantization can be very challenging, following the practice of existing literature \cite{lsq}, we quantize the first and last layer with 8-bit. During \acrshort{QAT}, we use per-tensor quantization \cite{krishnamoorthi} and learn the quantization scaling factor using backpropagation \cite{lsq} with a learning rate of 0.0001 in \acrshort{ADAM} optimizer. Our \acrshort{QAT} starts from a pre-trained full-precision network and is performed for 100 epochs. For all our \acrshort{QAT} experiments, we use \exma{} decay rate of 0.9999. In \qc{}, we train using \acrshort{ADAM} optimizer with a learning rate of 0.0001 to learn the correction scale factors and shift factors. We train correction factors for a single epoch while keeping the \acrfull{BN} statistics fixed. Rest of hyperparameters are used as default based on official \yoloa{}\footnote{\yoloa{}:  \href{https://github.com/ultralytics/yolov5}{https://github.com/ultralytics/yolov5}} and \yolob{}\footnote{\yolob{}: \href{https://github.com/WongKinYiu/yolov7}{https://github.com/WongKinYiu/yolov7}} implementations. Since, both \acrshort{LSQ} and Oscillation dampening perform experiments only on \imagenet{}, we reimplemented their methods for the task of object detection on \yolo{}. All our results are reported with standard object detection or semantic segmentation metric, namely mAP. Our code is in PyTorch and the experiments are performed on NVIDIA A-40 GPUs.

\subsection{Results on \yolo{} based object detection}

We evaluate both of our \exma{} and \qc{} techniques using \acrshort{LSQ}~\cite{lsq} on \yoloa{} and \yolob{} variants on \coco{} dataset for object detection task. We present results at different levels of precision \ie 3-bit, 4-bit, and 4-bit with even the first and last layer quantized to 4-bit. The object detection of quantized \yoloa{} and \yolob{} networks obtained by our proposed methods and their full precision (FP32) training are reported in~\tabref{tab:yolo-qat-ours}.

\input{tables/baselines-compare}

Our \qc{} method just by performing a post-hoc correction step consistently improves the \exma{} significantly for all different network architectures at 4-bit and 3-bit quantization. The improvement are especially significant on most efficient variants namely \yoloa{-n} and \yoloa{-s} at 3-bit ($\approx4-6\%$) and 4-bit ($\approx2\%$). Furthermore, even in the case of full quantization where even the first and last layers are quantized, our 4-bit quantization results using \qc{} consistently improve on our \acrshort{QAT} models trained with \exma{}. This clearly shows that latent weights that are stuck along the quantization thresholds can still be very useful if the error induced by those weights can be corrected using correction scale factors and shift factors learnt in our \qc{} method. Overall, our combined \exma{} and \qc{} method can reduce the gap between full precision models and 4-bit quantized models for all \yoloa{} and \yolob{} variants with a margin of around $\le 2.5\%$.


\subsection{Comparison against baselines}
\input{tables/perchannel-compare}

\input{tables/segment-yolo}

We also perform evaluation comparisons of our \exma{} and \qc{} techniques using \acrshort{LSQ} \cite{lsq} on \yoloa{} and \yolob{} variants on \coco{} dataset for object detection task against baseline methods namely, LSQ~\cite{lsq} and Oscillation dampening~\cite{nagel2022overcoming} at 4-bit and 3-bit quantization using \yoloa{} and \yolob{} on \coco{} dataset. Both \acrshort{LSQ}~\cite{lsq} and Oscillation dampening do not perform experiments on object detection and \yolo{} networks, so we re-implement their methods to create the baselines following their papers. The comparisons are done using the mAP metric and the results are reported in \tabref{tab:compare-baselines}. Both our \exma{} and \qc{} methods outperform \acrshort{LSQ} consistently and the gap between \acrshort{LSQ} and \qc{} is significant on both \yoloa{} and \yolob{} architectures with a margin of $\approx2-3\%$ consistently. Our \exma{} models are either comparable or sometimes even better than Oscillation dampening, reflecting the efficacy of \exma{} in reducing the effect of oscillation, especially resulting from activation quantization as oscillation dampening does not account for oscillation issue in activation quantization. Furthermore, our \qc{} method increases the gap ($\approx 2-3\%$) even further between our method and Oscillation dampening across both \yoloa{} and \yolob{} variants at 3-bit and 4-bit quantization.

\subsection{Comparison against per-channel quantization}

As mentioned in \secref{sec:qat-correction}, \qc{} scale and shift factors can be folded either in the succeeding \acrfull{BN} layer after the convolution layer or into quantization scale factors by converting per-tensor quantization to per-channel quantization. It has been noted previously that per-channel quantization tends to be more unstable \cite{whitepaper} for efficient networks with depth-wise convolutions due to a single scale factor being learnt for a depth-wise convolution filter (for eg. with size $3 \times 3$). Therefore, we further also provide experimental comparisons of our \qc{} method against per-channel quantization to reflect the efficacy of our method in improving the stability of per-channel \acrshort{QAT} by choosing \qc{} as a post-hoc correction step after \acrshort{QAT}. For this evaluation, we perform \acrshort{QAT} with \exma{} using per-tensor and per-channel quantization. We perform \qc{} only in case of per-tensor quantization and report the results in \tabref{tab:perchannel-compare}. First of all, as previous studies also noted, it can be observed that per-channel quantization with depth-wise convolutions can sometimes be inferior to per-tensor quantization. Furthermore, our \qc{} method on per-tensor quantization consistently produces better performance on both \yoloa{} and \yolob{} at 3-bit as well as 4-bit quantization with a margin of $\approx 3-4\%$ on \yolob and $\approx 2-4\%$ on \yoloa{} variants.

\vspace{1ex}

\subsection{Results on \yolo{} based semantic segmentation}
\vspace{0.5ex}
We further also evaluate our methods to quantize \yoloa{} variants at 3-bit and 4-bit on semantic segmentation task. We perform these experiments using \coco{} dataset and present results with mAP metric for the box and segmentation mask in \tabref{tab:segment-yolo}. Similar to the observations in the object detection task, our \qc{} method consistently improves the \exma{} method with a margin of $\approx 1-3 \%$ across \yoloa{-n} and \yoloa{-s} variants quantized at 3-bit and 4-bit. Furthermore, \qc{} in combination with \exma{} reduces the gap between full precision counterparts consistently on both detection box and segmentation mask metrics.

%% file: tables/yolo-qat-ours2.tex

\begin{table*}[]
\centering
\small
\caption{\em Our quantization-aware training performance using mAP metric for object detection task on the \coco dataset. * denotes first and last layers are trained at 4-bit quantization.}
\label{tab:yolo-qat-ours}
\vspace{-1ex}

\begin{tabular}{llcrrr|zzz}
\toprule
\multirow{2}{*}{Network} & \multirow{2}{*}{\# Params} & \multirow{2}{*}{FP} &  \multicolumn{3}{c|}{Ours (\exma)}                                                                 & \multicolumn{3}{z}{Ours (\exma + \qc)}                                                      \\
\cmidrule{4-9}
                                  &                                     &                                          & \multicolumn{1}{c}{4-bit} & \multicolumn{1}{c}{3-bit} & \multicolumn{1}{c|}{4-bit*} & \multicolumn{1}{z}{4-bit} & \multicolumn{1}{z}{3-bit} & \multicolumn{1}{z}{4-bit*} \\
                                  \midrule
\yoloa-n    & 1.87M                      & 28.0                & 22.1                      & 16.3                      & 16.5                        & \textbf{23.8}         & \textbf{18.2}         & \textbf{20.4}        \\
\yoloa-s    & 7.23M                      & 37.4                & 33.1                      & 28.5                      & 25.6                        & \textbf{34.0}         & \textbf{30.2}         & \textbf{32.0}        \\
\yoloa-m    & 21.2M                      & 45.2                & 42.1                      & 38.5                      & 38.5                        & \textbf{42.8}         & \textbf{40.0}         & \textbf{40.1}        \\
\yoloa-l    & 46.6M                      & 49.0                & 45.9                      & 43.1                      & 38.0                        & \textbf{46.6}         & \textbf{44.0}         & \textbf{43.6}        \\
\yoloa-x    & 86.7M                      & 50.7                & 47.8                      & 45.9                      & 40.6                        & \textbf{47.9}         & \textbf{46.8}         & \textbf{45.2}        \\
\yolob-tiny & 6.23M                      & 37.5                & 34.6                      & 30.3                      & 32.8                        & \textbf{35.2}         & \textbf{31.0}         & \textbf{34.3}        \\
\yolob      & 37.6M                      & 51.2                & 48.7                      & 46.2                      & 46.3                        & \textbf{48.9}         & \textbf{46.8}         & \textbf{47.6}        \\
\bottomrule
\end{tabular}
\end{table*}

%% file: tables/baselines-compare.tex
\begin{table}
\centering
\small
\caption{\em Comparison between \acrshort{LSQ}~\cite{lsq}, Oscillation dampening~\cite{nagel2022overcoming}, and our proposed method for quantization-aware training using mAP metric for object detection on \coco dataset.}
\vspace{-1ex}
\label{tab:compare-baselines}
\setlength{\tabcolsep}{1pt}
\begin{tabular}{llccc}
\toprule
Method                & \#-bit                  & \yoloa-n & \yoloa-s & \yolob-tiny \\
\midrule
Full-Precision        & 32-bit                  & 28.0      & 37.4    & 37.5       \\
\midrule
\acrshort{LSQ}~\cite{lsq}                  & \multirow{4}{*}{4-bit} & 20.6   & 32.4   & 32.9      \\
Osc. Damp.~\cite{nagel2022overcoming} &                         & 21.5   & 32.9  & 33.5         \\
Ours (\exma)            &                         & 22.1    & 33.1    & 34.6       \\
\rowcolor{myGray}
Ours (\exma+\qc)         &                         & \textbf{23.8}    & \textbf{34.0}      & \textbf{35.2}       \\
\midrule
\acrshort{LSQ}~\cite{lsq}                  & \multirow{4}{*}{3-bit} & 15.2    & 27.2    & 28.4      \\
Osc. Damp.~\cite{nagel2022overcoming} &                         & 16.4   & 27.5   & 29.2          \\
Ours (\exma)            &                         & 16.4    & 28.5    & 30.3       \\
\rowcolor{myGray}
Ours (\exma+\qc)         &                         & \textbf{18.2}    & \textbf{30.2}    & \textbf{31.0}        \\
\bottomrule
\end{tabular}
\end{table}

%% file: tables/perchannel-compare.tex

\begin{table}[t]
\centering
\small
\caption{\em Comparison between per-channel quantization against our \qc~method. We train using \exma~ and \acrshort{LSQ} at different bit-width on \coco dataset. Note, * denotes first and last layers are also trained at 4-bit quantization.}
\vspace{-0.5ex}
\label{tab:perchannel-compare}
\setlength{\tabcolsep}{1.5pt}

\begin{tabular}{lllccz}
\toprule
\multirow{2}{*}{Network}    & \multirow{2}{*}{\# Params} & \multirow{2}{*}{\#-bits} & \multicolumn{3}{c}{\acrshort{LSQ} + Ours (\exma)} \\
\cmidrule{4-6}
                            &                            &                          & Per-tensor & Per-channel & Ours (\qc) \\
\midrule
                            
\multirow{3}{*}{\yoloa-n}   & \multirow{3}{*}{1.87M}    &  4-bit                    & 22.1       & 22.1        & \textbf{23.8}      \\
                            &                            & 3-bit                    & 16.3       & 14.4        & \textbf{18.2}      \\
                            &                            & 4-bit*                   & 16.5       & 19.4        & \textbf{20.4}      \\
                            \midrule

\multirow{3}{*}{\yoloa-s}   & \multirow{3}{*}{7.23M}      & 4-bit                    & 33.1       & 32.6        & \textbf{34.0}        \\
                            &                            & 3-bit                    & 28.5       & 27.3        & \textbf{30.2}      \\
                            &                            & 4-bit*                   & 25.6       & 31.2        & \textbf{32.0}        \\
                            \midrule

\multirow{3}{*}{\yolob-tiny} & \multirow{3}{*}{6.23M}                             & 4-bit                    & 34.6       & 32.3        & \textbf{35.2}      \\
                            &                            & 3-bit                    & 30.3       & 27.3        & \textbf{31.0}        \\
                            &                            & 4-bit*                   & 32.8       & 30.3        & \textbf{34.3}     \\
                            \bottomrule
\end{tabular}
\end{table}

%% file: tables/segment-yolo.tex
\begin{table}[h]
\centering
\small

\caption{\em Comparison with baseline method \ie, \acrshort{LSQ}~\cite{lsq} against our proposed methods using mAP metric for semantic segmentation task on the COCO dataset. * denotes first and last layers are trained at 4-bit quantization. Our methods consistently outperforms baseline methods on both \yoloa-{n,s} variants at 3-bit and 4-bit quantization.}
\vspace{-0.5ex}
\setlength{\tabcolsep}{1.2pt}
\label{tab:segment-yolo}
\begin{tabular}{llcccc}
\toprule
\multirow{2}{*}{Method} & \multirow{2}{*}{\#-bit} & \multicolumn{2}{c}{Mask (mAP)}                            & \multicolumn{2}{c}{Box (mAP)}                             \\
\cmidrule{3-6}
                        &                         & \multicolumn{1}{l}{\yoloa-n} & \multicolumn{1}{l}{\yoloa-s} & \multicolumn{1}{l}{\yoloa-n} & \multicolumn{1}{l}{\yoloa-s} \\
                        \midrule
Full-Precision          & 32-bit                  & 23.4          & 31.7         & 27.6         & 37.6         \\
\midrule
Baseline                & \multirow{3}{*}{4-bit}  & 16.5          & 27.8         & 18.5         & 32.1         \\
Ours (\exma)              &                         & 17.7          & 28.3         & 19.8         & 32.6         \\
\cellcolor{myGray} Ours (\exma+\qc)           &                         &  \cellcolor{myGray} \textbf{19.5}          & \cellcolor{myGray} \textbf{29.4}         & \cellcolor{myGray} \textbf{22.3}         & \cellcolor{myGray} \textbf{33.7}         \\
\midrule
Baseline                & \multirow{3}{*}{3-bit}  & 12.5          & 23.9         & 14.1         & 27.1         \\
Ours (\exma)              &                         & 13.9          & 24.7         & 15.9         & 27.8         \\
\cellcolor{myGray} Ours (\exma+\qc)           &                         & \cellcolor{myGray} \textbf{15.8}          & \cellcolor{myGray} \textbf{25.5}         & \cellcolor{myGray} \textbf{18.1}         & \cellcolor{myGray} \textbf{29.6}         \\
\midrule
Baseline                & \multirow{3}{*}{4-bit*} & 15.7          & 26.0           & 17.1         & 30.0           \\
Ours (\exma)              &                         & 16.5          & 26.7         & 17.9         & 30.4         \\
\cellcolor{myGray} Ours (\exma+\qc)           &                         &  \cellcolor{myGray} \textbf{18.3}          & \cellcolor{myGray} \textbf{27.2}         & \cellcolor{myGray} \textbf{20.8}         & \cellcolor{myGray} \textbf{31.5}
 
 \\
\bottomrule
\end{tabular}
\vspace{-2ex}
\end{table}

%% file: text/conclusion.tex
\vspace{2ex}
\section{Discussion}
\vspace{0.5ex}

In this work, we perform the first study for \acrshort{QAT} on efficient real-time \yoloa{}, \yolob{} detectors and show that these networks suffer from oscillation issue. We further show that the oscillation issue does not only affect weight quantization but also activation quantization on \yolo{} models. To mitigate side-effects of oscillations due to \acrshort{STE} approximation of rounding function and per-tensor quantization, we introduce two simple techniques, namely \exma{} and \qc{}. Our proposed \acrshort{QAT} pipeline combining \exma{} and \qc{} produces a new state-of-the-art quantized \yolo{} models at low-bit precision (3-bits and 4-bits). In future work, we believe \qc{} scale and shift factors can be generalized by estimating correction factors that are weighted for specific regions in the tensors that could potentially lead to even further performance gains.

%% file: text/appendix.tex
Here, we provide additional experimental results and analysis. First, we provide a comparison between vanilla baseline methods and results using \qc{} on the baselines on \yoloa{} and \yolob{}. Later, we also provide ablation studies of various setups of \qc{} and show the stability of \exma{} for the decay parameter $(\alpha)$. 

\section{Comparisons with vanilla baselines and \qc{} based baselines}
\input{tables/baseline-qc}
Further, we also perform experiments to evaluate the effectiveness of \qc{} on the baseline \acrshort{QAT} methods such as \acrshort{LSQ}~\cite{lsq} and Oscillation dampening~\cite{nagel2022overcoming} on object detection task on \coco{} dataset and the results are reported in \tabref{tab:baseline-qc}. Our \qc{} approach to correct the error induced due to oscillating weights and scale factors cannot only improve the detection performance of quantized models of \exma{} but also the baseline methods. Despite that, our combined approach with both \exma{} and \qc{} outperforms all the baselines with \qc{} consistently at 4-bit as well as 3-bit quantization on \yoloa{} and \yolob{} variants.

\section{Ablation on different \qc{} setups}
\input{tables/qc-ablation}
\qc{} can correct the error induced due to oscillations after the quantization by employing the per-channel scale and shift correction factors. These scale factors can also be chosen per-tensor. To evaluate the effectiveness of different components of \qc{} such as scale and shift correction factors, we provide ablation studies in \tabref{tab:qc-ablation}. We also provide results on both per-tensor and per-channel setup of \qc{}. It is important to note that both scale and shift parameters are equally important in both per-tensor and per-channel \qc{} setup and neither alone can effectively reduce oscillation-based error. Also, even the simple per-tensor setup of \qc{} improves the \exma{} performance but as expected it cannot meet the performance achieved by the per-channel \qc{} setting.

\section{Effect of varying decay factor in \exma{}}
\input{tables/diff-ema-momentum}
To check the stability of \exma{} to varying decay factors, we trained different 4-bit \yoloa{-n} models using \coco{} datasets and the comparisons are provided in \tabref{tab:diff-mom-ema}. As shown, our \exma{} approach is quite stable to different decay parameters. \exma{} takes into into account $\approx1-(1-\alpha)$ iterations to compute the average weights or scale factors. Typically, oscillations are consistent over $\ge100$ iterations, and taking an average of scale factors or weights over $\ge 100$ iterations works effectively in mitigating the oscillation side-effects in the final \acrshort{QAT} model. 

\section{Oscillation in scale factors with or without \exma{}}
\begin{figure*}[ht]
     \centering
     \begin{subfigure}[b]{0.48\textwidth}
         \centering
         \includegraphics[width=0.99\textwidth]{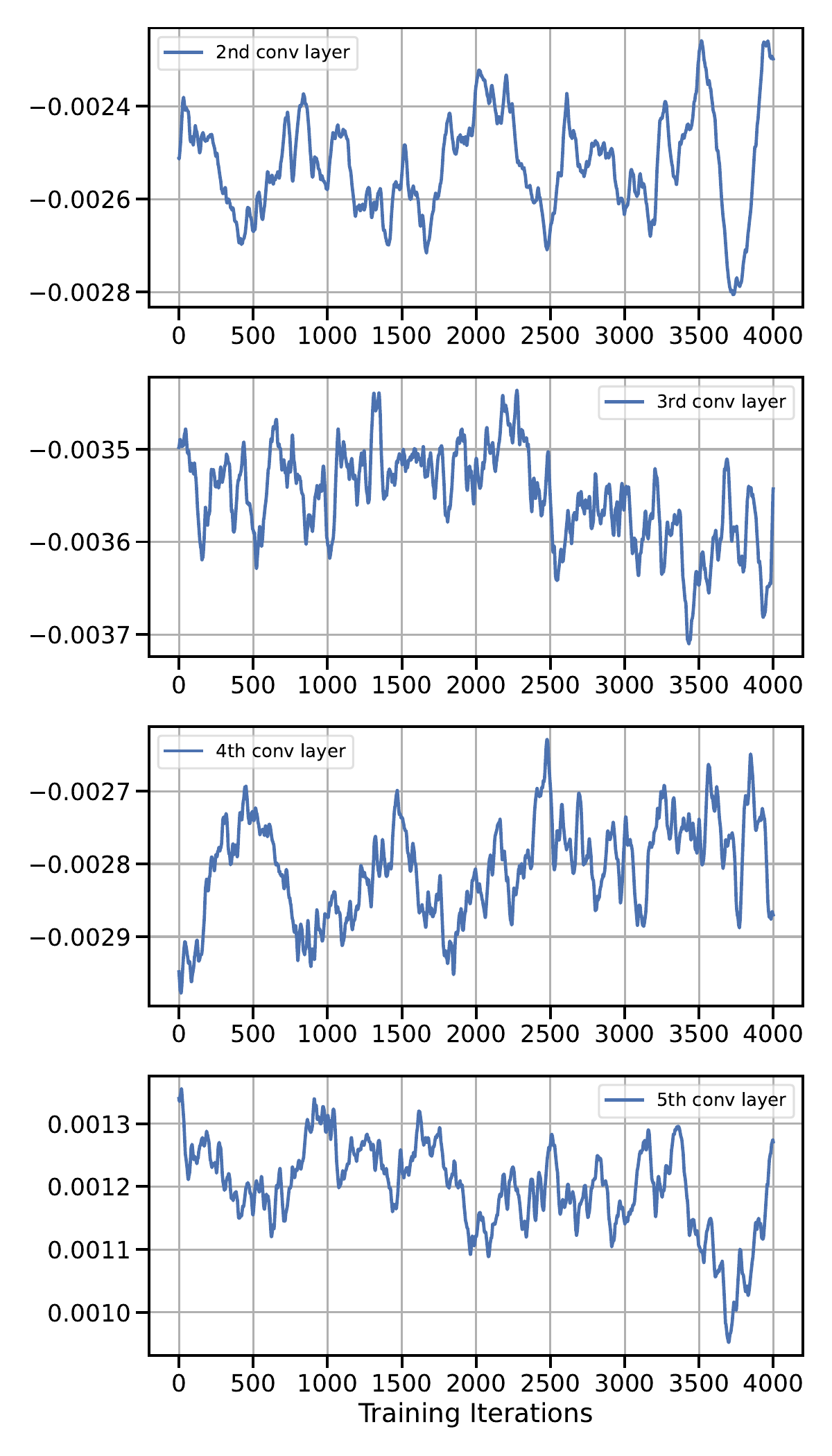}
         \caption{}
     \end{subfigure}
     \begin{subfigure}[b]{0.48\textwidth}
         \centering
         \includegraphics[width=0.99\textwidth]{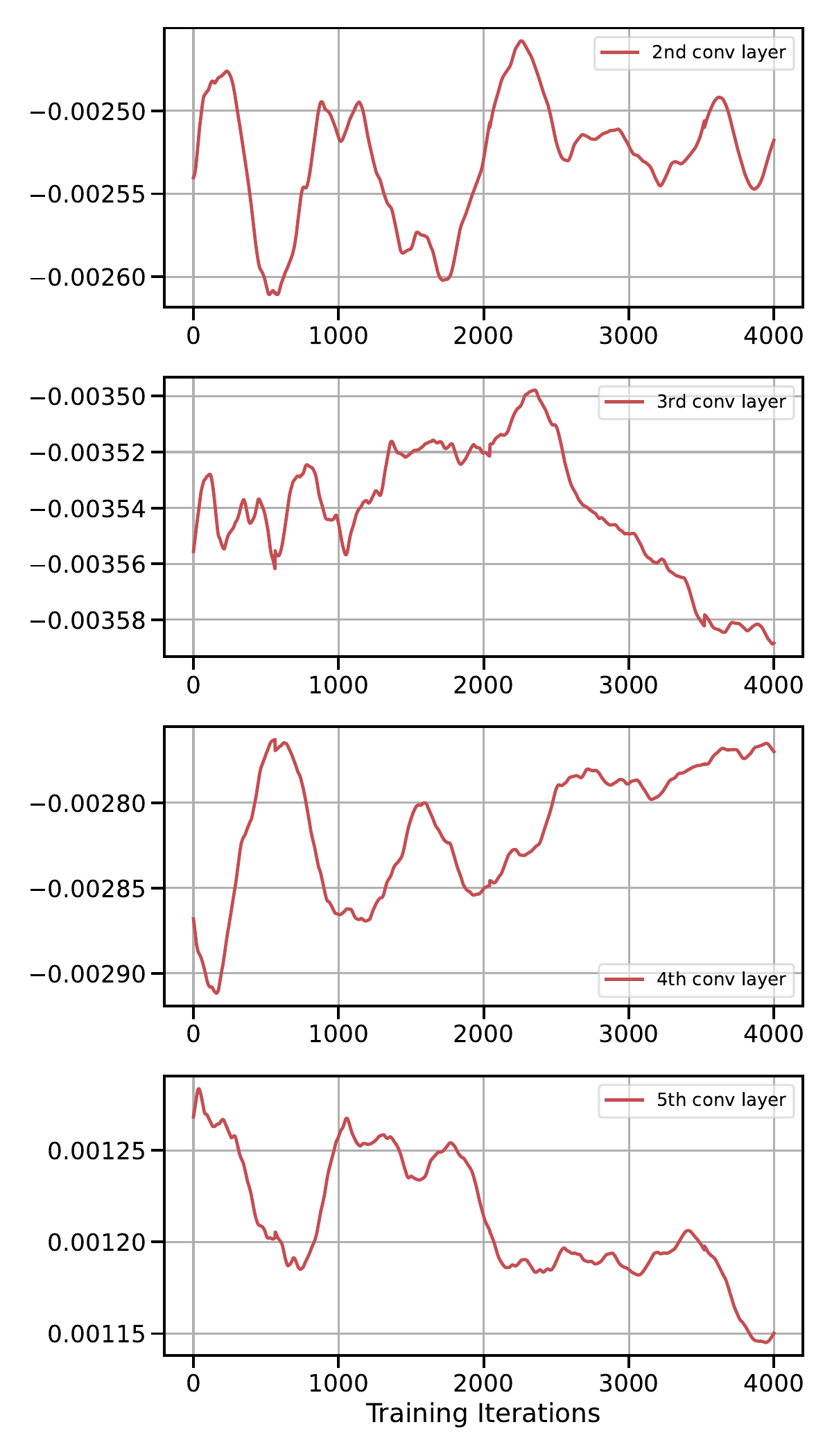}
         \caption{}
     \end{subfigure}
        \caption{\em Effect of \exma{} on oscillation issue in \yoloa{}-n variant trained on \coco{} dataset at 4-bit precision using \acrshort{LSQ}~\cite{lsq}. (a) Scale factors for weight quantization in the vanilla model during the last 4K iterations of 2nd-5th conv layer, (b) Scale factors for weight quantization in \exma{} model during the last 4K iterations of 2nd-5th conv layer. Here, it can be observed that \exma{} makes the quantization scale factors stable for all the layers and gets rid of the oscillation issue.}
        \label{fig:scale-wt-yolov5}
\end{figure*}

\begin{figure*}[ht]
     \centering
     \begin{subfigure}[b]{0.48\textwidth}
         \centering
         \includegraphics[width=0.99\textwidth]{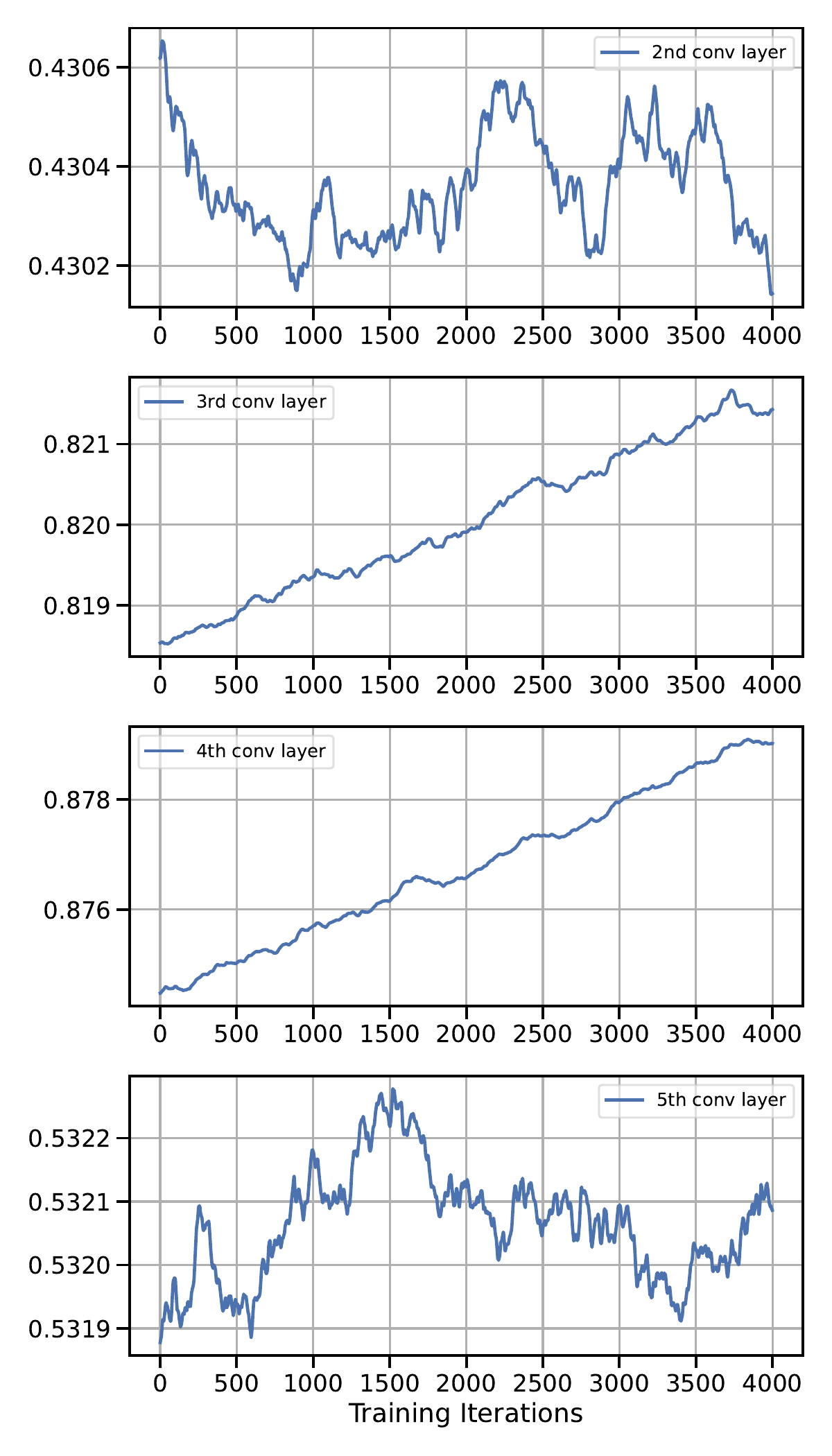}
         \caption{}
     \end{subfigure}
     \begin{subfigure}[b]{0.48\textwidth}
         \centering
         \includegraphics[width=0.99\textwidth]{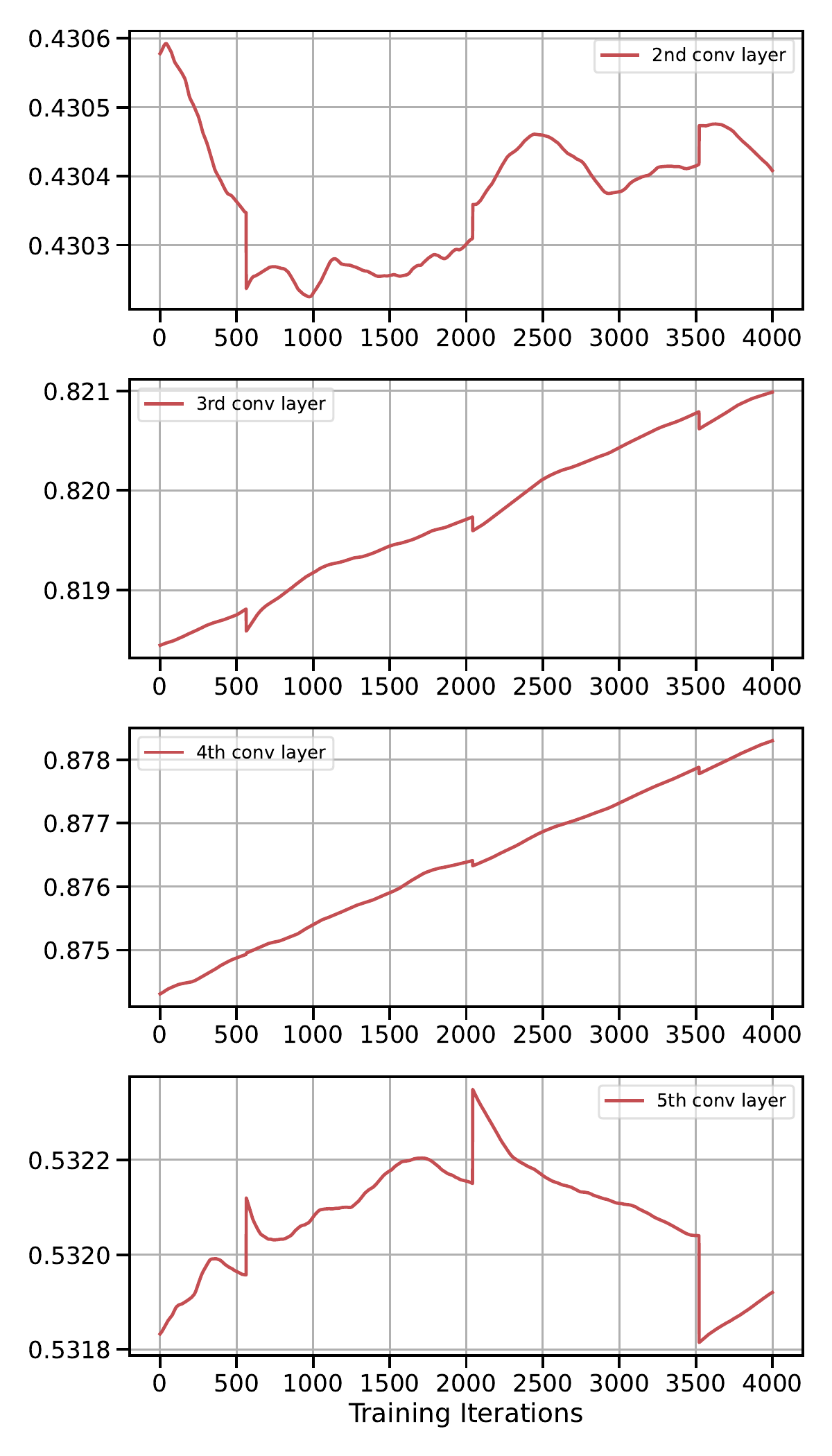}
         \caption{}
     \end{subfigure}
        \caption{\em Effect of \exma{} on oscillation issue in \yoloa{}-n variant trained on \coco{} dataset at 4-bit precision using \acrshort{LSQ}~\cite{lsq}. (a) Scale factors for activation quantization in the vanilla model during the last 4K iterations of 2nd-5th conv layer, (b) Scale factors for activation quantization in \exma{} model during the last 4K iterations of 2nd-5th conv layer. Here, it can be observed that \exma{} makes the quantization scale factors stable for all the layers and gets rid of the oscillation issue.}
        \label{fig:scale-act-yolov5}
\end{figure*}

To show the effect of \exma{} on the quantization scale factors during the training, we also provide the plots for scale factors during the last 4K iterations of training with or without \exma{} in \figref{fig:scale-wt-yolov5} and \figref{fig:scale-act-yolov5} for scale factors of weights and activations respectively. It can be seen that \exma{} leads to a smoother transition of quantization scale factors for both weights and activations throughout the training and thus lead to stable training of quantized \yolo{} models.

%% file: tables/baseline-qc.tex
\begin{table}[h]

\centering
\caption{\em Comparison between \acrshort{LSQ}~\cite{lsq}, Oscillation dampening~\cite{nagel2022overcoming}, and our proposed method after performing our \qc{} for quantization-aware training using mAP metric for object detection task on the \coco dataset.}
\vspace{1ex}
\label{tab:baseline-qc}
\begin{tabular}{lclccc}
\toprule
Method                                 & Ours (\qc{}) & \#-bit                 & \yoloa{-n} & \yoloa{-s} & \yolob{-tiny} \\
\midrule
Full-Precision                         & -         & 32-bit                 & 28.0      & 37.4    & 37.5       \\
\midrule
\multirow{2}{*}{\acrshort{LSQ}~\cite{lsq}}                   & \xmark         & \multirow{6}{*}{4-bit} & 20.6    & 32.4    & 32.9       \\
                                       & \cmark         &                        & 22.6    & 33.3    & 34.1       \\
                                       \cmidrule{4-6}
\multirow{2}{*}{Osc. Dampening~\cite{nagel2022overcoming}} & \xmark         &                        & 21.5    & 32.9    & 33.5       \\
                                       & \cmark         &                        & 23.1    & 33.4    & 34.3       \\
                                       \cmidrule{4-6}
\multirow{2}{*}{Ours (\exma)}            & \xmark         &                        & 22.1    & 33.1    & 34.6       \\
                                       & \cmark         &                        &  \cellcolor{myGray} \textbf{23.8}    & \cellcolor{myGray} \textbf{34.0}      & \cellcolor{myGray} \textbf{35.2}       \\
\midrule
\multirow{2}{*}{\acrshort{LSQ}~\cite{lsq}}                   & \xmark         & \multirow{6}{*}{3-bit} & 15.2    & 27.2    & 28.4       \\
                                       & \cmark         &                        & 17.1    & 29.4    & 30.2       \\
\cmidrule{4-6}
\multirow{2}{*}{Osc. Dampening~\cite{nagel2022overcoming}} & \xmark         &                        & 16.4    & 27.5    & 29.2       \\
                                       & \cmark         &                        & 17.9    & 29.6    & 30.5       \\
                                       \cmidrule{4-6}

\multirow{2}{*}{Ours (\exma)}            & \xmark         &                        & 16.4    & 28.5    & 30.3       \\
                                       & \cmark         &                        & \cellcolor{myGray} \textbf{18.2}    & \cellcolor{myGray} \textbf{30.2}    & \cellcolor{myGray} \textbf{31.0}   \\
\bottomrule
\end{tabular}
\end{table}

%% file: tables/qc-ablation.tex
\begin{table}[]
\centering
\caption{\em Ablation studies of different \qc{} setups, where either per-tensor or per-channel correction is performed varying whether to use \qc{} scale or shift on \yoloa{-n} trained at 4-bit on \coco{} dataset.}
\vspace{1ex}
\label{tab:qc-ablation}
\begin{tabular}{lccr}
\toprule
\qc{} Setup                     & \qc{} Scale & \qc{} Shift & mAP  \\
\midrule
\multirow{3}{*}{Per-tensor}  & \cmark        & \xmark        & 22.2 \\
                             & \xmark        & \cmark        & 22.3 \\
                             & \cmark        & \cmark        & 22.6 \\
                             \midrule
\multirow{3}{*}{Per-channel} & \cmark        & \xmark        & 23.5 \\
                             & \xmark        & \cmark        & 23.6 \\
                             & \cellcolor{myGray} \cmark        & \cellcolor{myGray} \cmark        & \cellcolor{myGray} \textbf{23.8} \\
\bottomrule
\end{tabular}
\end{table}

%% file: tables/diff-ema-momentum.tex
\begin{table}[]
\centering
\caption{\em Different values of decay parameter $(\alpha)$ in \exma{} for \yoloa{-n} trained at 4-bit on \coco{} dataset. Note, \exma{} is quite stable with respect to different decay parameters. }
\vspace{1ex}
\label{tab:diff-mom-ema}
\begin{tabular}{cr}
\toprule
Decay parameter $(\alpha)$ & mAP  \\
\midrule
0.0              & 20.6 \\
0.9            & 21.5 \\
\cellcolor{myGray} 0.99           & \cellcolor{myGray}\textbf{22.1} \\
\cellcolor{myGray} 0.999          & \cellcolor{myGray} \textbf{22.1} \\
\cellcolor{myGray} 0.9999         & \cellcolor{myGray} \textbf{22.1} \\
\bottomrule
\end{tabular}
\end{table}